\pdfoutput=1

\documentclass[10pt,twocolumn,letterpaper]{article}

\usepackage{cvpr}
\usepackage{times}
\usepackage{epsfig}
\usepackage{graphicx}
\usepackage{amsmath}
\usepackage{amssymb}

\usepackage[ruled,norelsize]{algorithm2e}
\usepackage{algorithmic}
\usepackage{multirow}
\usepackage{tabularx}

\usepackage[breaklinks=true,bookmarks=false]{hyperref}

\cvprfinalcopy 

\def\cvprPaperID{23} 

\begin{document}

\title{DisplaceNet: Recognising Displaced People from Images by Exploiting Dominance Level}

\author{
	Grigorios Kalliatakis, \,\,
	Shoaib Ehsan, \,\,
	Maria Fasli, \,\,
	Klaus McDonald-Maier\vspace{3pt}\\
	School of Computer Science and Electronic Engineering, University of Essex\\
	{\tt\small \{gkallia, sehsan, mfasli, kdm\}@essex.ac.uk} 
}

\maketitle

\begin{abstract}
	Every year millions of men, women and children are forced to leave their homes and seek refuge from wars, human rights violations, persecution, and natural disasters. The number of forcibly displaced people came at a record rate of 44,400 every day throughout 2017, raising the cumulative total to 68.5 million at the year’s end, overtaken the total population of the United Kingdom. Up to 85\% of the forcibly displaced find refuge in low- and middle-income countries, calling for increased humanitarian assistance worldwide. To reduce the amount of manual labour required for human-rights-related image analysis, we introduce DisplaceNet, a novel model
	which infers potential displaced people from images by integrating the control level of the situation and conventional convolutional neural network (CNN) classifier into one framework for image classification. Experimental results show that DisplaceNet achieves up to 4\% coverage\textendash the proportion of a data set for which a classifier is able to produce a prediction\textendash gain over the sole use of a CNN classifier. Our dataset, codes and trained models will be available online at \url{https://github.com/GKalliatakis/DisplaceNet}
\end{abstract}

\section{Introduction}

The displacement of people refers to the forced movement of people from their locality or environment and occupational activities \footnote{A distinction is often made between conflict-induced and disaster-induced displacement, yet the lines between them may be blurred in practice.}. It is a form of social change caused by a number of factors such as armed conflict, violence, persecution and human rights violations. Globally, there are now almost 68.5 million forcibly displaced people\textendash and most are hosted in developing regions, while today 1 out of every 110 people in the world is displaced \cite{global_trends_report}.

\begin{figure}[t!]
	\centering
	\begin{tabular}{cc}
		\includegraphics[width=0.2\textwidth,height=0.38\textheight,keepaspectratio]{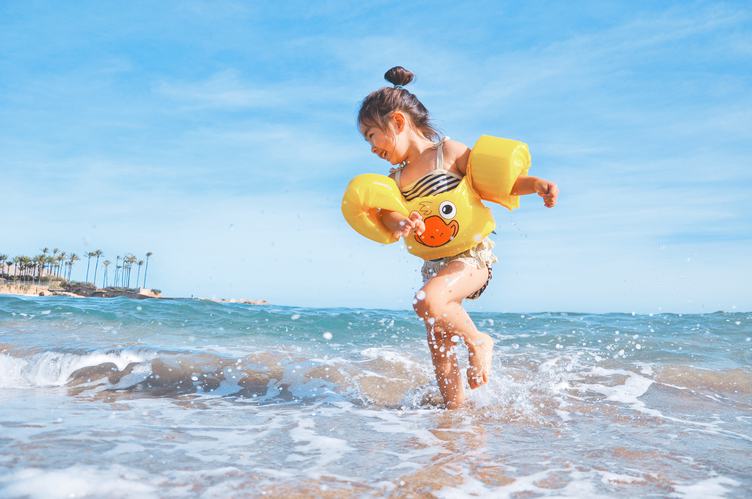} &   \includegraphics[width=0.22\textwidth,height=0.43\textheight,keepaspectratio]{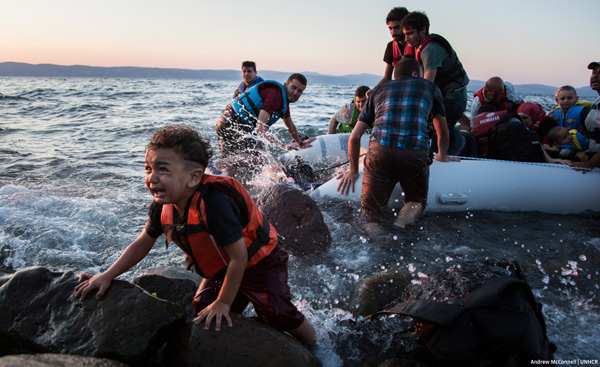} \\
		(a) Child playing & (b) Displaced people \\[7pt]
	\end{tabular}
	\caption{Displaced people recognition poses a challenge at a higher level for the well-studied, deep image representation learning methods. Regularly, emotional states can be the only notifying difference between the encoded visual content of an image that depicts a non-violent situation and the encoded visual content of an image displaying displaced people.}
	\label{Fig. 1}
\end{figure}

In the era of social media and big data, the use of visual evidence to document conflict and human rights abuse has become an important element for human rights organizations and advocates. However, the omnipresence of visual evidence
may deluge those accountable for analysing it. Currently, information extraction from human-rights-related imagery requires manual labour by human rights analysts and advocates. Such analysis is time consuming, expensive, and remains emotionally traumatic for analysts to focus on images of horrific events. In this work, we strive to reconcile this gap by automating parts of this process; given a single image we label the image as either \textit{displaced people} or \textit{non displaced people}. Figure \ref{Fig. 1} illustrates that naive schemes based solely on object detection or scene recognition are doomed to
fail in this binary classification problem. If we can exploit existing smartphone cameras, which are ubiquitous, it may be possible to turn recognition of displaced populations into a powerful and cost-effective computer vision application that could improve humanitarian responses.

Recently, Kalliatakis \textit{et al.} \cite{kalliatakis2019exploring} shown that a two-stage fine-tuning of deep Convolutional Neural Networks (CNNs) can address the multi-class classification problem of human rights violations to a certain extent. In this paper, we introduce \textit{DisplaceNet}, a novel method designed with a human-centric approach for solving a sought-after, binary classification problem in the context of human rights image analysis; \textit{displaced people recognition}. Our hypothesis is that the control level of the situation by the person, ranging from \textit{submissive} / \textit{non-control} to \textit{dominant} / \textit{in-control}, is a powerful cue that can help our network make a distinction between displaced people and non-violent instances. First, we develop an end-to-end model for recognising rich information about people's emotional states by jointly analysing the person and the whole scene. We use the continuous dimensions of the \textit{VAD Emotional State Model} \cite{mehrabian1995framework}, which describes emotions using three numerical dimensions: Valence (V); Arousal (A); and Dominance (D). In the context of this work, we have focused only on dominance\textendash measures the control level of the situation by the person\textendash because it is considered as the most relevant for the task of recognising displaced people. Second, following the estimation of emotional states, we introduce a new method for interpreting the overall dominance level of an entire image sample based on the emotional states of all individuals on the scene. As a final step, we propose to assign weights to image samples according to the image-to-overall-dominance relevance to guide prediction of the image classifier. 

We carried out comprehensive experimentation to evaluate our method for displaced people recognition on a subset of HRA dataset \cite{kalliatakis2019exploring}. This subset contains all image samples from the \textit{displaced people} category (positive samples) alongside the same number of images taken from the \textit{no violation} category (negative samples). Experimental results
show that DisplaceNet can improve the coverage\textendash the proportion of a data set for which a classifier is able to produce a prediction\textendash by 4\% compared to the sole use of a CNN classifier that is trained end-to-end using the same training data.

\section{Related Work}

\noindent
\textbf{Human rights image analysis.} Image analysis in the context of human rights plays a crucial role in human rights advocacy and accountability efforts. Automatic perception of potential human rights violations enables scientists and investigators to discover content, that may otherwise be concealed by sheer volume of visual data. The automated systems concerned with human rights abuses identification are not producing evidence, but are instead narrowing down the volume of material that must be examined by human analysts who are making legitimate claims, that they then present in justice, accountability, or advocacy settings \cite{aronson2018computer}. There is a considerable body on literature for video analysis with respect to human rights \cite{piracés_2018,aronson2015video}. A different group of methods based on still images alongside the first ever publicly available image dataset for the purpose of human rights violations recognition was introduced in \cite{visapp17}. Recently, Kalliatakis \textit{et al.} \cite{kalliatakis2019exploring} introduced a larger, verified-by-experts image dataset for fine-tuning object-centric and scene-centric CNNs.

\noindent
\textbf{Object detection.} On of the most improved areas of computer vision in the past few years is object detection, the process of determining the instance of the class to which an object belongs and estimating the location of the object. Object detectors can be split into two main categories: one-stage detectors and two-stage detectors. One of the first modern one-stage object detectors based on deep networks is OverFeat \cite{OverFeat}, which applies a sliding window approach based on multi-scaling for jointly performing classification, detection and localization. More recent works such as YOLO \cite{redmon2016you,redmon2016yolo9000} and SSD \cite{Fu2017DSSDD,liu2016ssd}, have revived interest in one-stage methods, mainly due to their real time efficiency. The leading model in modern object detectors is based on a two-stage approach which was established in \cite{uijlings2013selective}. R-CNN, a notably successful family of methods, \cite{girshick2015fast,girshick2014rich} enhanced the second-stage classifier to a convolutional network, resulting in large accuracy improvements. After that, the speed of R-CNN has also improved over the years by integrating region proposal networks (RPN) with the second-stage classifier into a single convolution network, known as the Faster R-CNN framework \cite{ren2015faster}. Our method belongs to one-stage detectors. Specifically, we adopt the RetinaNet framework \cite{lin2018focal} that handles class imbalance by reshaping the standard cross entropy loss to focus training on a sparse set of hard examples and down-weights the loss assigned to well-classified examples. 

\noindent
\textbf{Emotion recognition.} Most of the research in computer vision to recognise people’s emotional states is focused on facial expression analysis \cite{fabian2016emotionet,eleftheriadis2016joint} where a large variety of methods have been developed to recognise the 6 basic emotions defined by Ekman and Friesen \cite{ekman1971constants}. Lately, CNNs have been used as backbone for the facial expression recognition of Action Units \cite{fabian2016emotionet}. The second family of methods for emotion recognition use the continuous dimensions of the \textit{VAD Emotional State Model} \cite{mehrabian1995framework,mehrabian1974approach} to represent emotions instead of discrete emotion categories. The VAD model uses a 3-dimensional approach to describe and measure the emotional experience of humans: Valence (V) describes affective states from highly negative (unpleasant) to highly positive (pleasant); Arousal (A) measures the intensity of affective states ranging from calm to excited or alert; and Dominance (D) represents the feeling of being controlled or influenced by external stimuli.

\section{Method}

\begin{figure}[t!]
	\centering
	
	\includegraphics[width=0.48\textwidth,height=0.40\textheight,keepaspectratio]{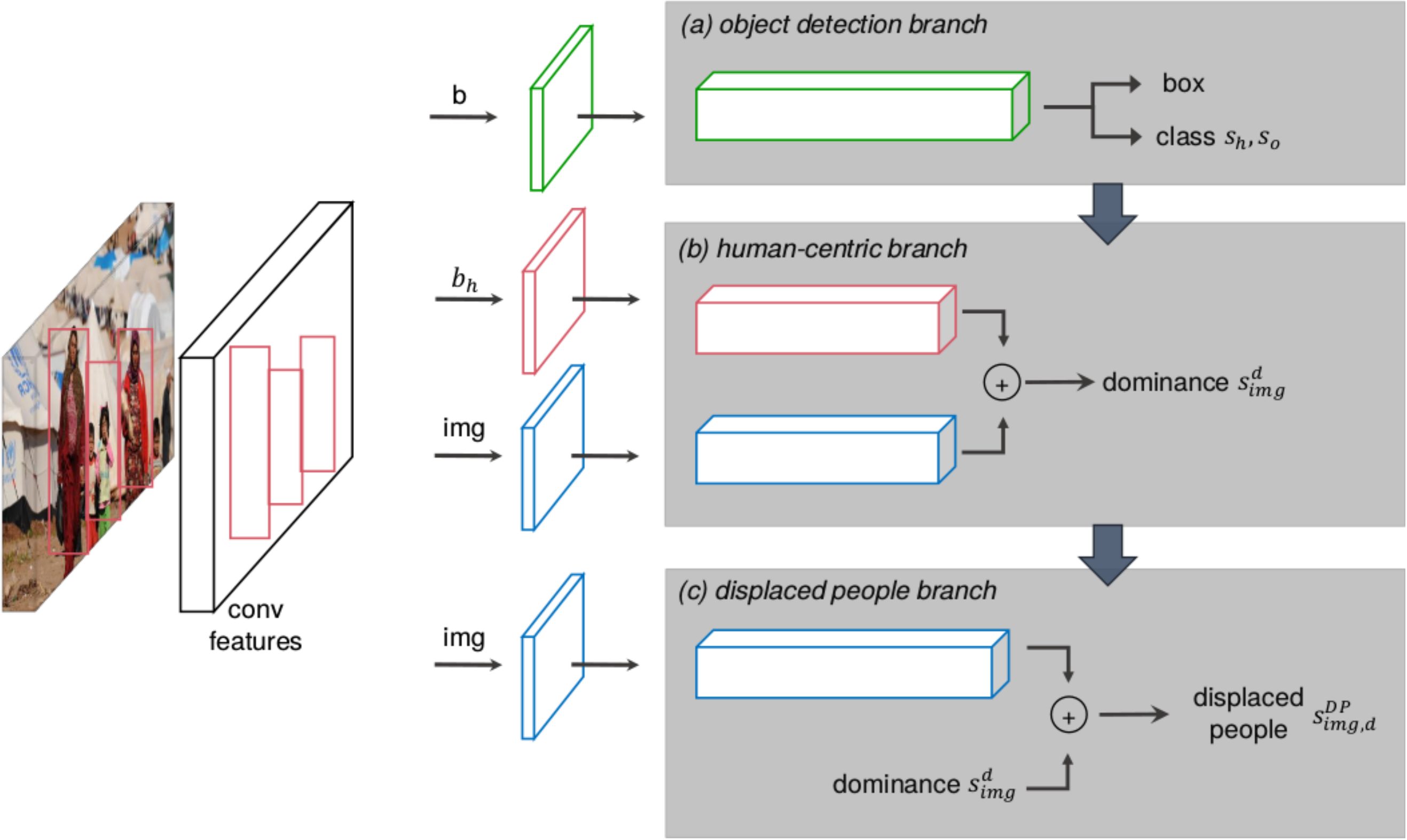} 
	\caption{Model Architecture. Our model consists of (a) an \textit{object detection} branch, (b) \textit{human-centric} branch, and (c) a \textit{displaced people} branch. The image features and their layers are shared between the human-centric and displaced people branches (blue boxes).}
	\label{Fig. 2}
\end{figure}

We now describe our method for recognising displaced people by exploiting the dominance level of the entire image. Our goal is to label challenging everyday photos as either `displaced populations' or `no displaced populations'.  

First, in order to detect the emotional traits of an image, we need to accurately localise the box containing a $human$ and the associated object of interaction (denoted by $b_h$ and $b_o$, respectively), as well as identify the emotional states $e$ of each human using the VAD model. Our proposed solution adopts the RetinaNet \cite{lin2018focal} object detection framework alongside an additional \textit{human-centric} branch that estimates the continuous dimensions of each detected person and then determines the overall dominance level of the given image. 

Specifically, given a set of candidate boxes, RetinaNet outputs a set of object boxes and a class label for each box. While the object detector can predict multiple class labels, our model is concerned only with the `person' class. The region of the image comprising the person whose feelings are to be estimated at $b_h$ is used alongside the entire image for simultaneously extracting their most relevant features. These features, are fused and used to perform continuous emotion recognition in VAD space. Our model extends typical image classification by assigning a triplet score $s_{img,d}^{DP}$  to pairs of candidate human boxes $b_h$ and the displaced people category \textit{a}. To do so, we decompose the triplet score into three terms:

\begin{equation} \label{eq:1}
s_{img,d}^{DP} = s_h  \cdot s_{h,img}^{d}  \cdot s_{img}^{DP}
\end{equation}

\noindent
We discuss each component next, followed by details for training and inference. The overall architecture of DisplaceNet is shown in
Figure \ref{Fig. 2}.

\subsection{Model components}

\noindent
\textbf{Object detection.} The object detection branch of DisplaceNet is identical to that of RetinaNet \cite{lin2018focal} single stage classifier. First, an image is forwarded through ResNet-50 \cite{he2016deep}, then in the subsequent pyramid layers, the more semantically important features are extracted and concatenated with the original features for improved bounding box regression. 

\noindent
\textbf{Human-centric branch.} The first role of the human-centric branch is to assign an emotion classification score to each human box . Similar to \cite{kosti2017emotion}, we use an end-to-end model with three main modules: two feature extractors and a fusion module. The first module takes the region of the image comprising the person whose emotional traits are to be estimated, $b_h$, while the second module takes as input the entire image and extracts global features. This way the required contextual support is accommodated in our emotion recognition process. Finally, the third module takes as input the extracted image and body features and estimates the continuous dimensions in VAD space. 

The second role of the human-centric branch is to assign a dominance score $s_{img}^{d}$ which characterises the entire input image.
$s_{img}^{d}$  is the encoding of the overall dominance score relative to human box $b_h$ and entire image $img$, that is:

\begin{equation} \label{eq:2}
s_{img}^{d} = \frac{1}{n} \sum_{i=1}^{n} s_{h,img}^d  
\end{equation} 

\begin{figure}[t!]
	\centering
	\begin{tabular}{cc}
		\includegraphics[width=0.23\textwidth,height=0.4\textheight,keepaspectratio]{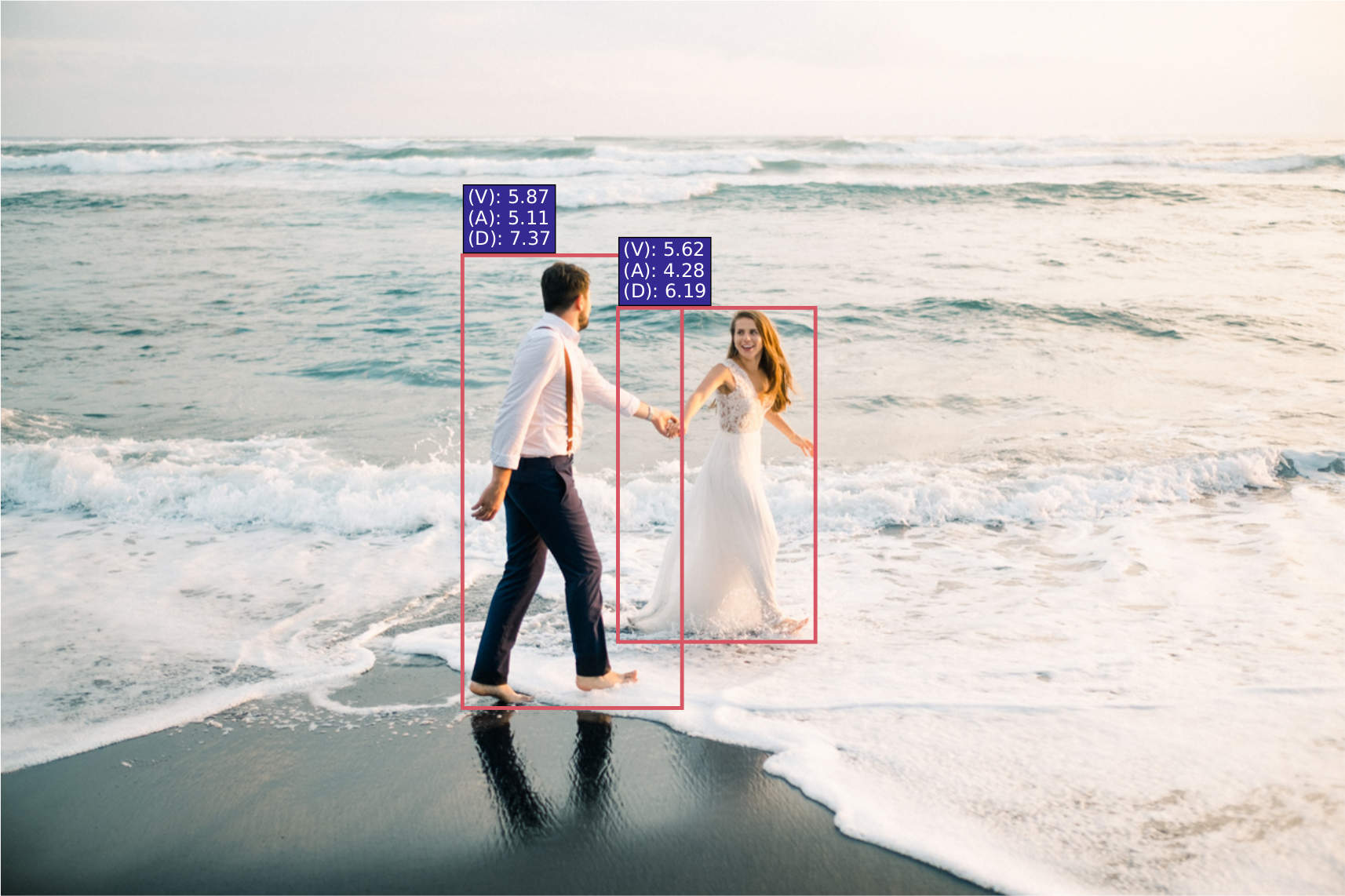} &   \includegraphics[width=0.23\textwidth,height=0.4\textheight,keepaspectratio]{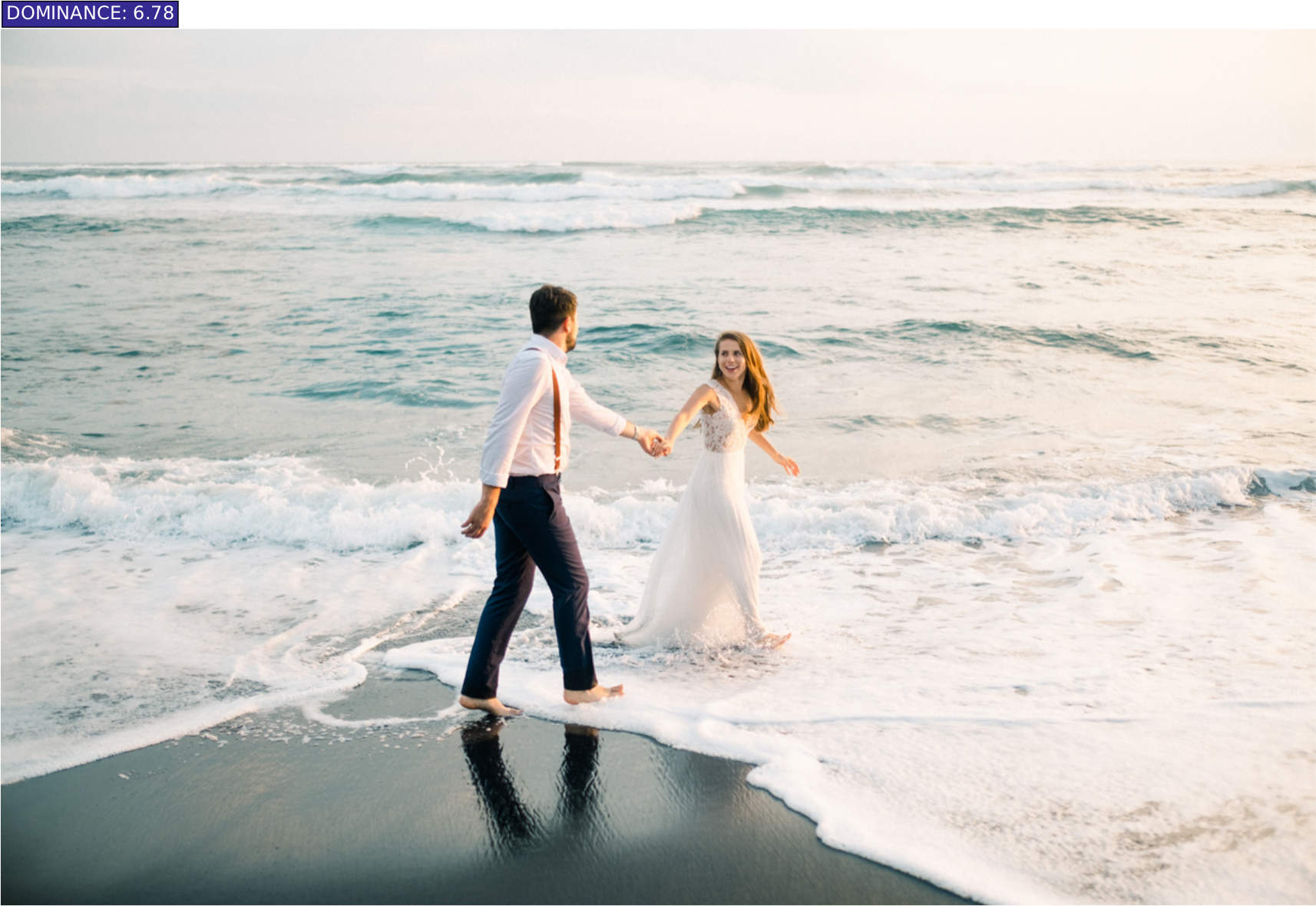} \\
		(a)  & (b)  \\[7pt]
		\includegraphics[width=0.23\textwidth,height=0.4\textheight,keepaspectratio]{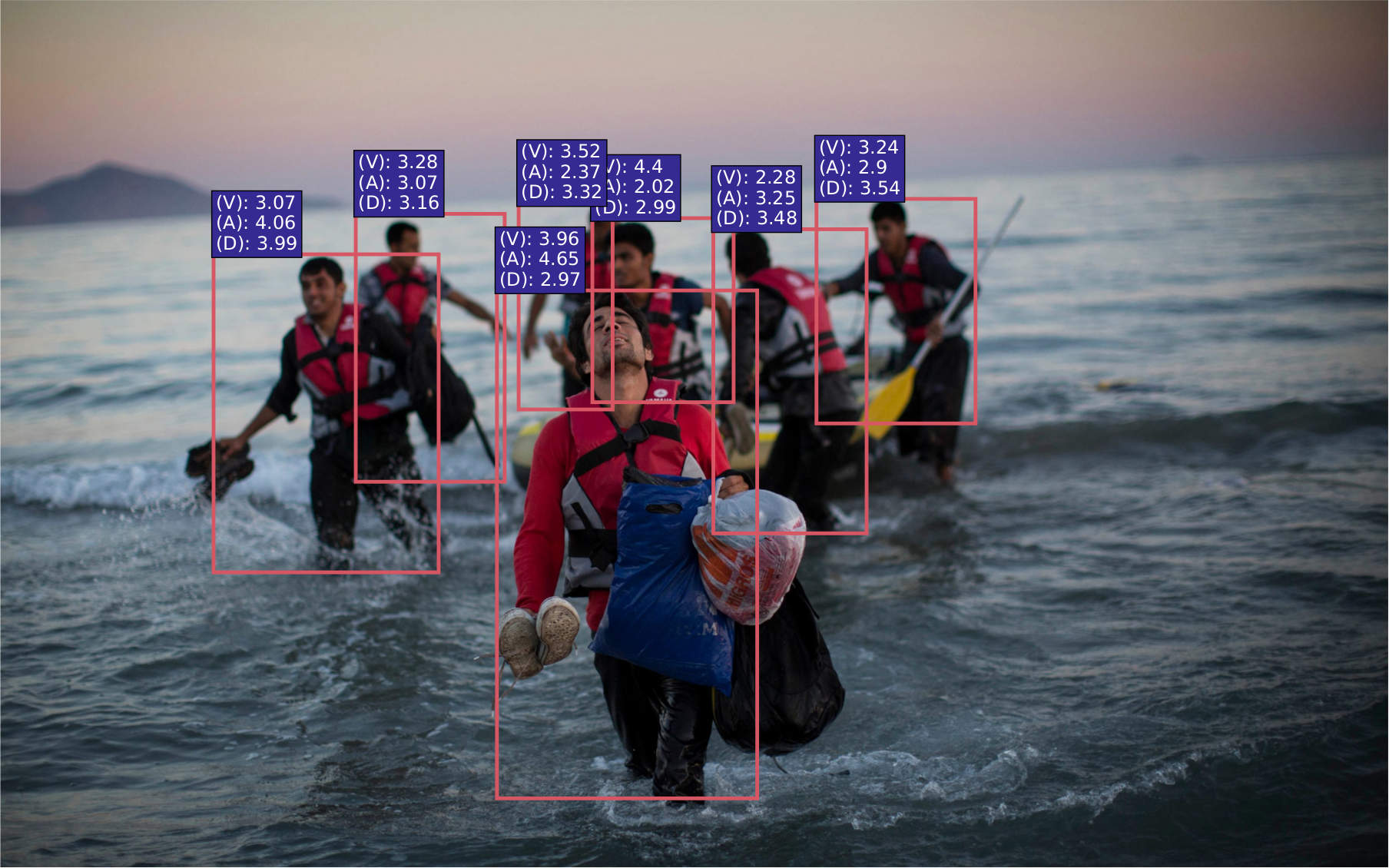} &   \includegraphics[width=0.23\textwidth,height=0.4\textheight,keepaspectratio]{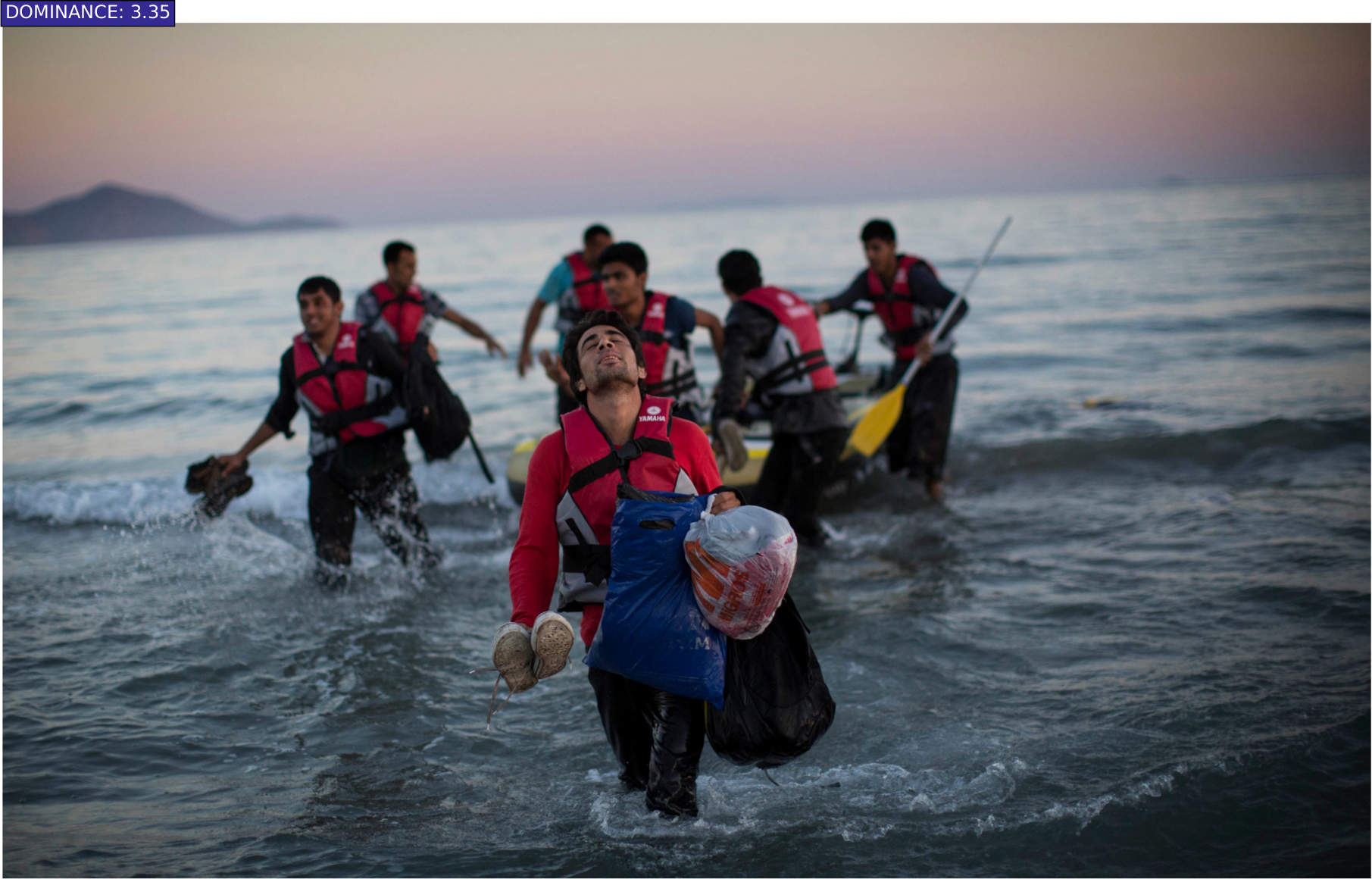} \\
		(c)  & (d)  \\[7pt]
	\end{tabular}
	\caption{Estimating continuous emotions in VAD space vs overall dominance from the combined body and image features. The left column shows the predicted emotional states and their scores from the person region of interest (RoI), while the right column show the same images analysed for overall dominance. The dominance score will be integrated with the standard image classification scores $s_{img}^{DP}$ to identify displaced people. }
	\label{Fig. 3}
\end{figure}

\noindent
Figure \ref{Fig. 3} (a),(c) illustrates the three different emotional states over the estimated target objects locations while Figure \ref{Fig. 3} (b),(d) shows the overall dominance score proposed here. Note that although all three predicted numerical dimensions are depicted, only dominance is considered to be the most relevant to the task of recognising displaced people since the other two dimensions can be ambiguous for several situations.

\noindent
\textbf{Displaced people recognition.} The first role of the displaced people branch is to assign a classification score to the input image. Similar to two-phase transfer learning scheme introduced in \cite{kalliatakis2019exploring}, we train an end-to-end model for binary classification of everyday photos as either `displaced populations' or `no displaced populations'. In order to improve the discriminative power of our model, the second role of the displaced people branch is to integrate $s_{img}^{d}$ in the recognition pipeline. Specifically, the raw image classification score  is readjusted based on the inferred dominance score. Each dominance unit, that is deltas from the neutral state, is expressed as a numeric weight varying between 1 and 10, while the neutral states of dominance are assigned between 4.5 and 5.5 based on the number of examples per each of the scores in the continuous dimensions reported in \cite{kosti2017emotion}.  The adjustment that will be assigned to the raw probability, $s_{img}^{DP}$ is the weight of dominance multiplied by a factor of 0.11 which has been experimentally set. When the input image depicts positive dominance, the adjustment factor is subtracted from the positive human-rights-abuse probability and added to the negative human-rights-abuse probability. Similarly, when the input image depicts negative dominance the adjustment factor is added to the negative human-rights-abuse probability and subtracted from the positive human-rights-abuse probability. This is formally written in Algorithm \ref{algorithm}.

Finally, when no $b_h$ is detected from the object detection branch, (\ref{eq:1}) is reduced into plain image classification as follows:

\begin{equation} \label{eq:6}
s_{img,d}^{DP} = s_{img}^{DP} 
\end{equation}

\begin{algorithm}[t!] 
	\caption{Calculate $s_{img,d}^{DP}$} 
	\label{algorithm} 
	\begin{algorithmic} 
		\REQUIRE $b_h > 0$
		\vskip 0.05in
		\STATE $s_{pos}\gets s_{img}^{dp}$ \COMMENT{$dp$: positive displaced people}
		\STATE $s_{neg}\gets s_{img}^{ndp}$ \COMMENT{$ndp$: negative displaced people}
		\vskip 0.05in
		\IF {$weight$ $\geq$ 4.5  \AND $weight$ $\leq$ 5.5} 
		\STATE Return $s_{pos}, s_{neg}$
		\ELSIF {$weight> 5.5$} 
		\STATE $diff = weight-5.5$
		\STATE $adj = diff*0.11$
		\STATE $s_{pos} = s_{pos}-adj$
		\STATE $s_{neg} = s_{neg}+adj$
		\ELSIF {$weight< 4.5$} 
		\STATE $diff = 4.5-weight$
		\STATE $adj = diff*0.11$
		\STATE $s_{pos} = s_{pos}+adj$
		\STATE $s_{neg} = s_{neg}-adj$
		\ENDIF
		\STATE Return $s_{pos}, s_{neg}$
	\end{algorithmic}
\end{algorithm}

\begin{figure*}[t!]
	\centering
	\begin{tabular}{ccc}
		\includegraphics[width=0.28\textwidth,height=0.3\textheight,keepaspectratio]{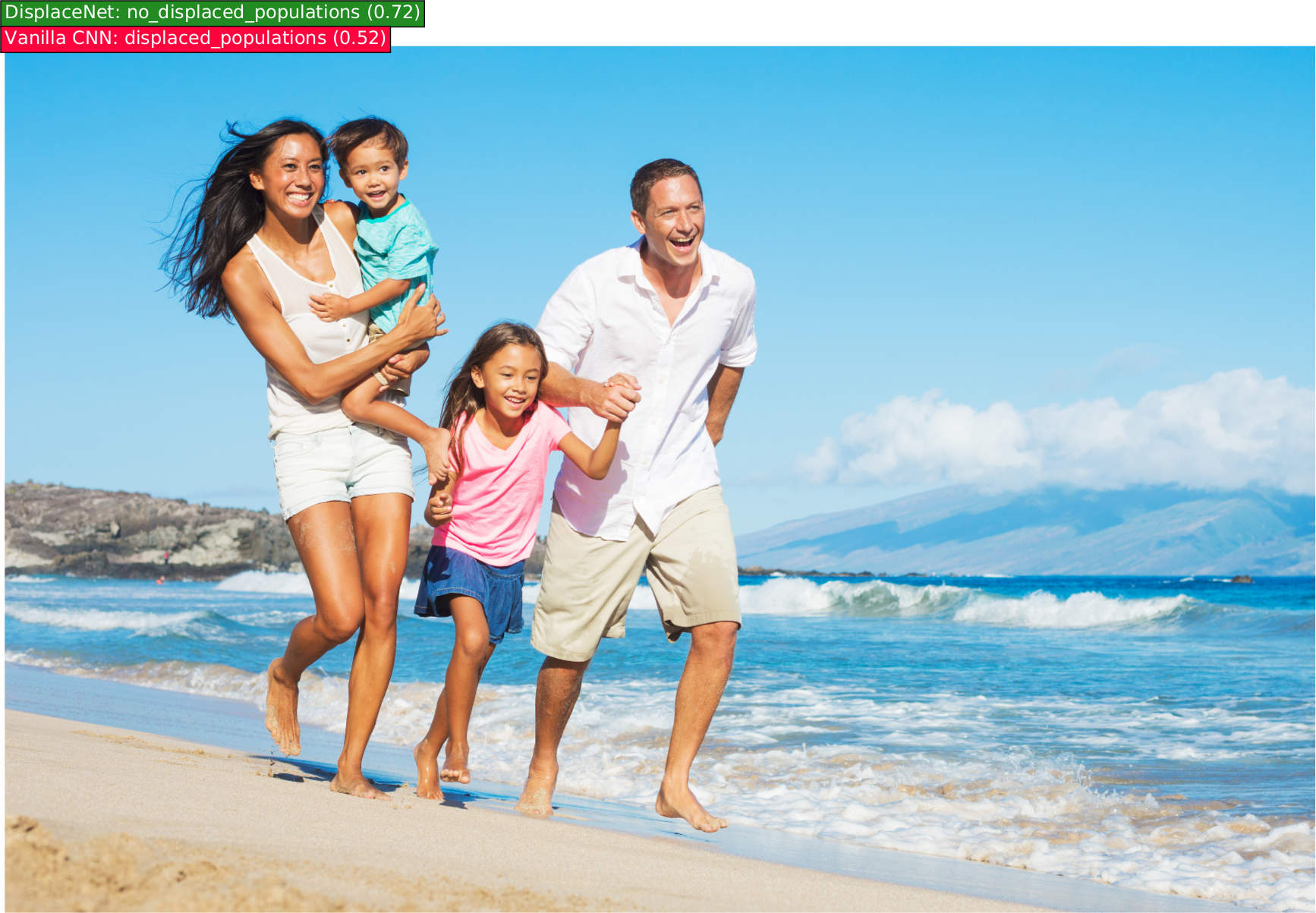} &   \includegraphics[width=0.28\textwidth,height=0.3\textheight,keepaspectratio]{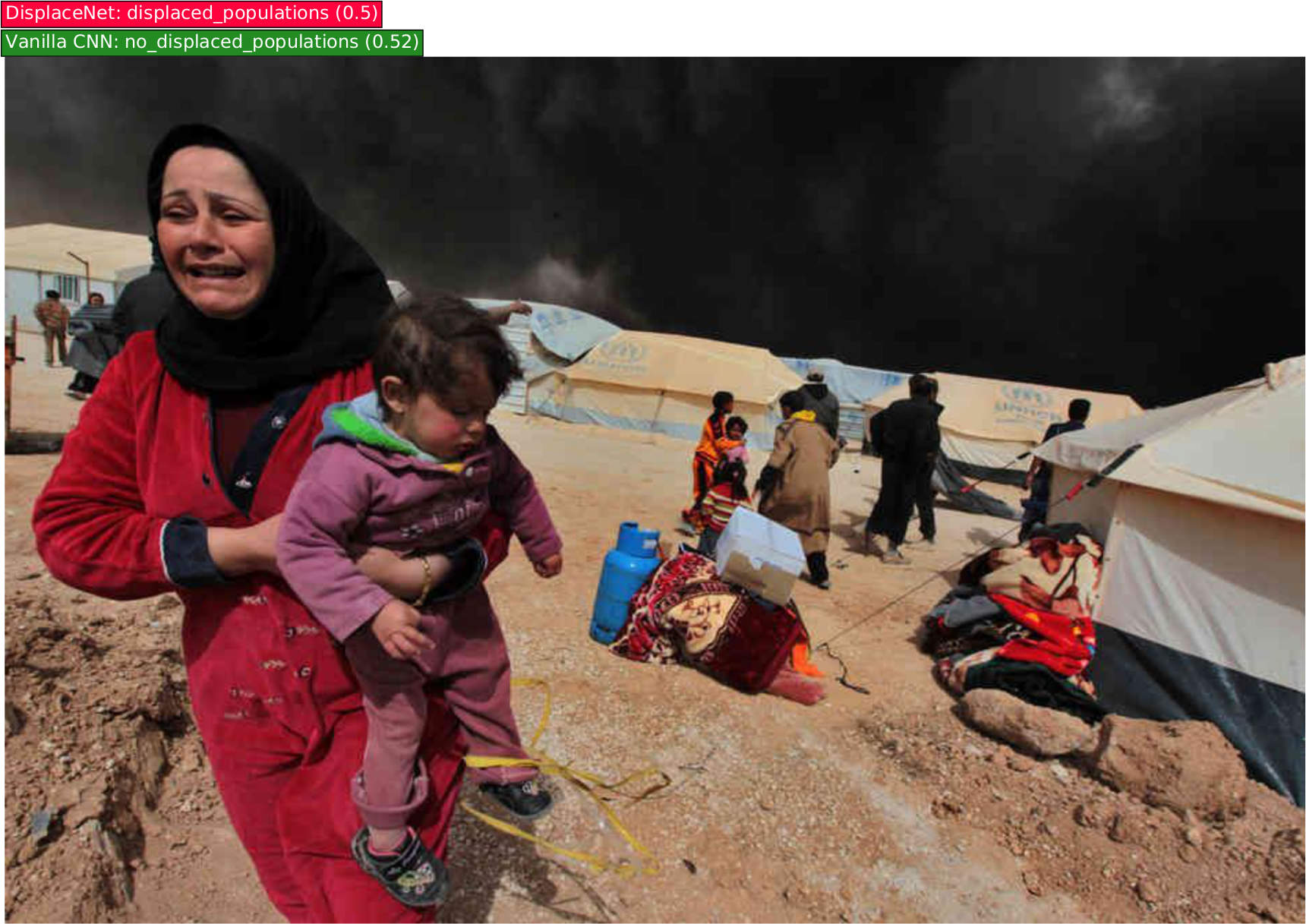} &
		\includegraphics[width=0.31\textwidth,height=0.4\textheight,keepaspectratio]{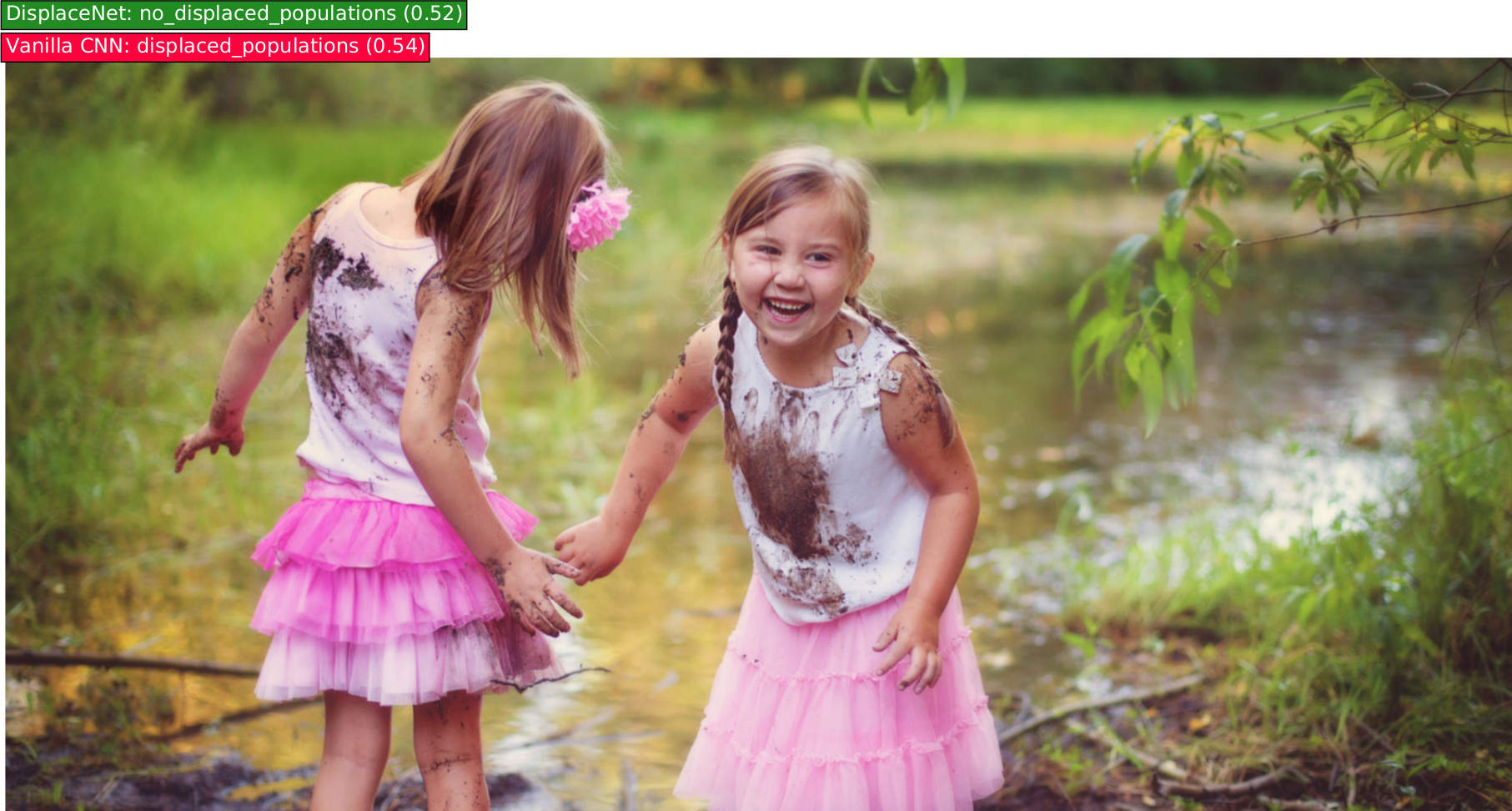}\\
		(a) & (b)  & (c) \\[-1pt]
		\includegraphics[width=0.29\textwidth,height=0.3\textheight,keepaspectratio]{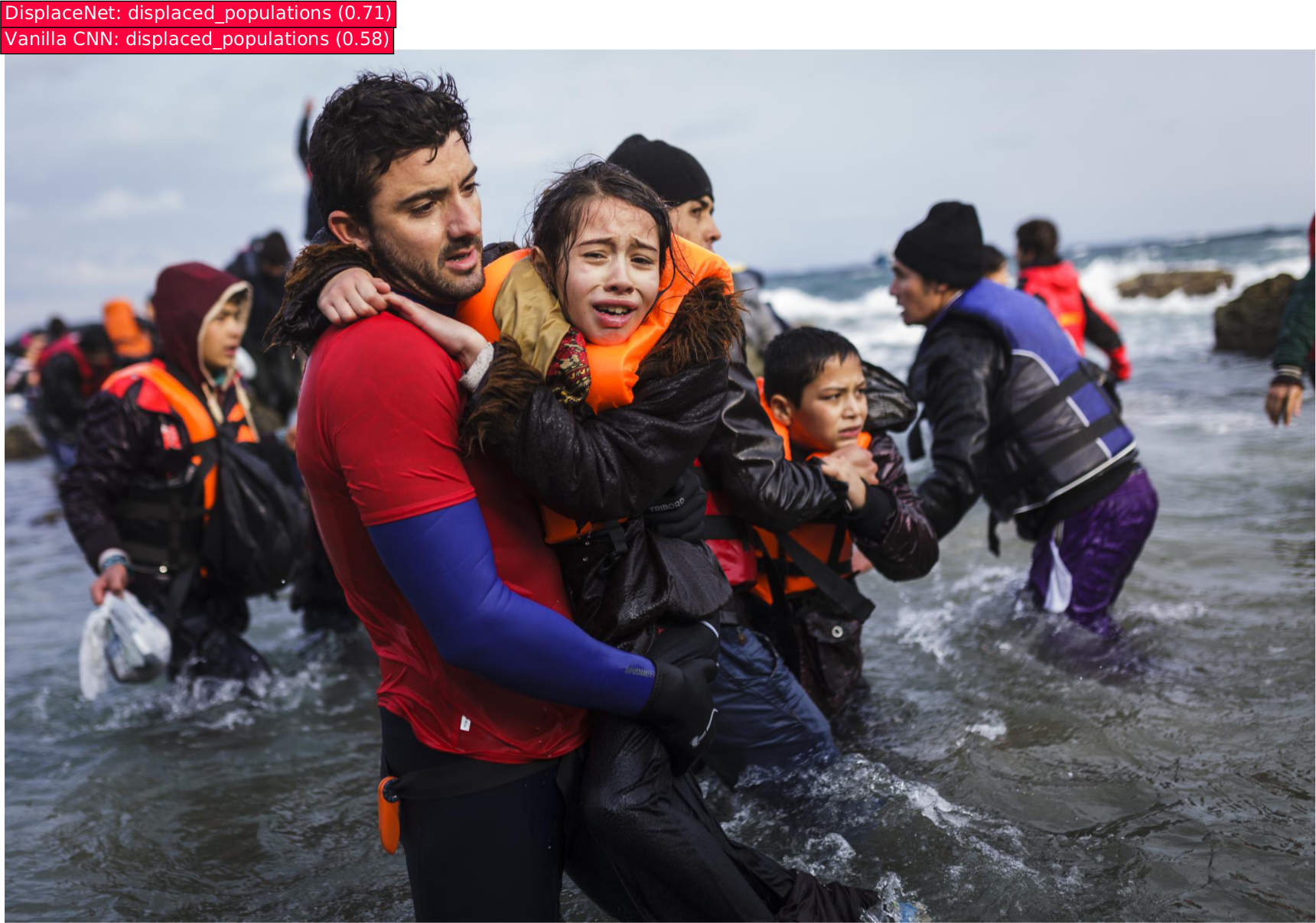} &   \includegraphics[width=0.35\textwidth,height=0.4\textheight,keepaspectratio]{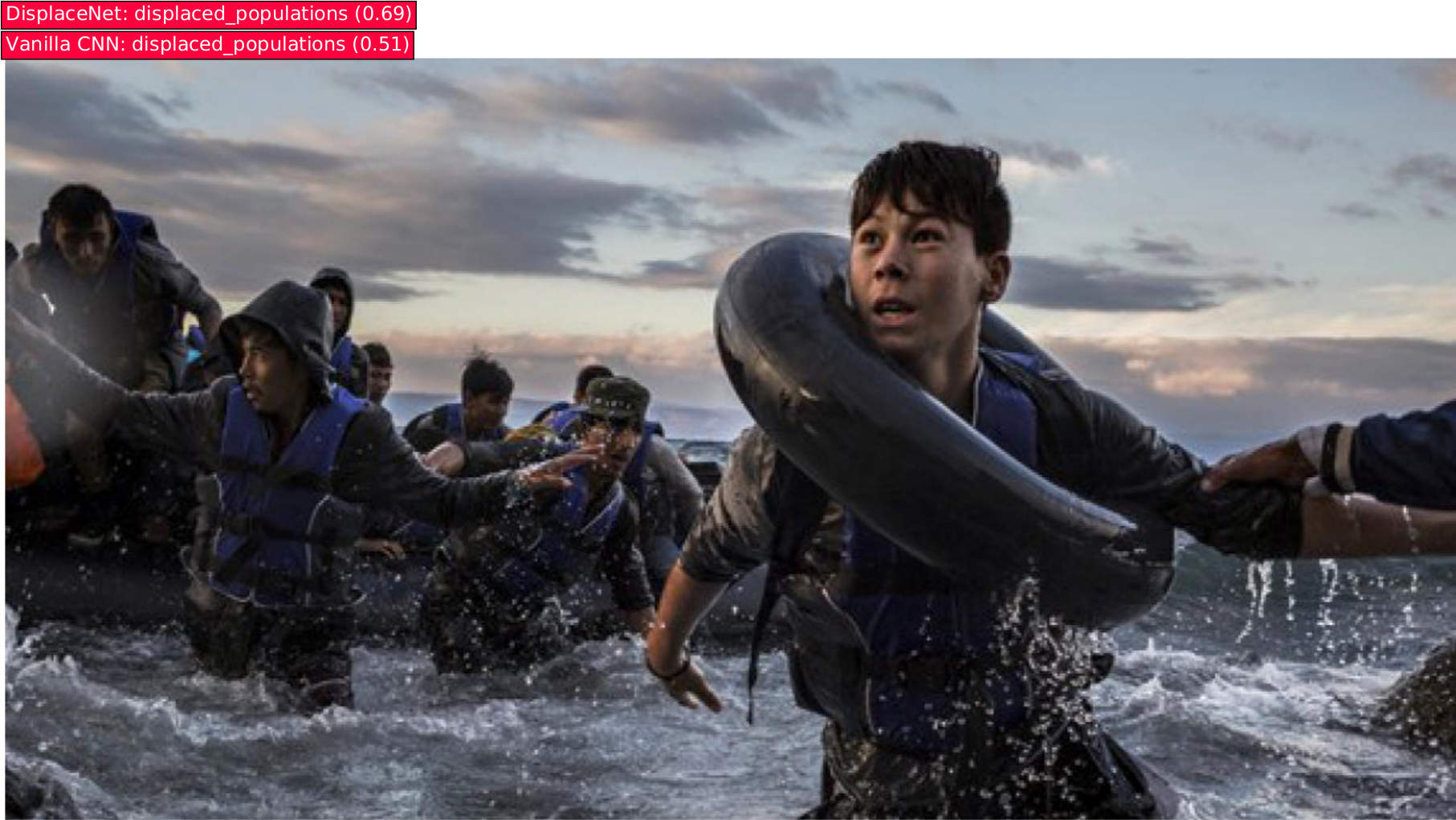} &
		\includegraphics[width=0.26\textwidth,height=0.3\textheight,keepaspectratio]{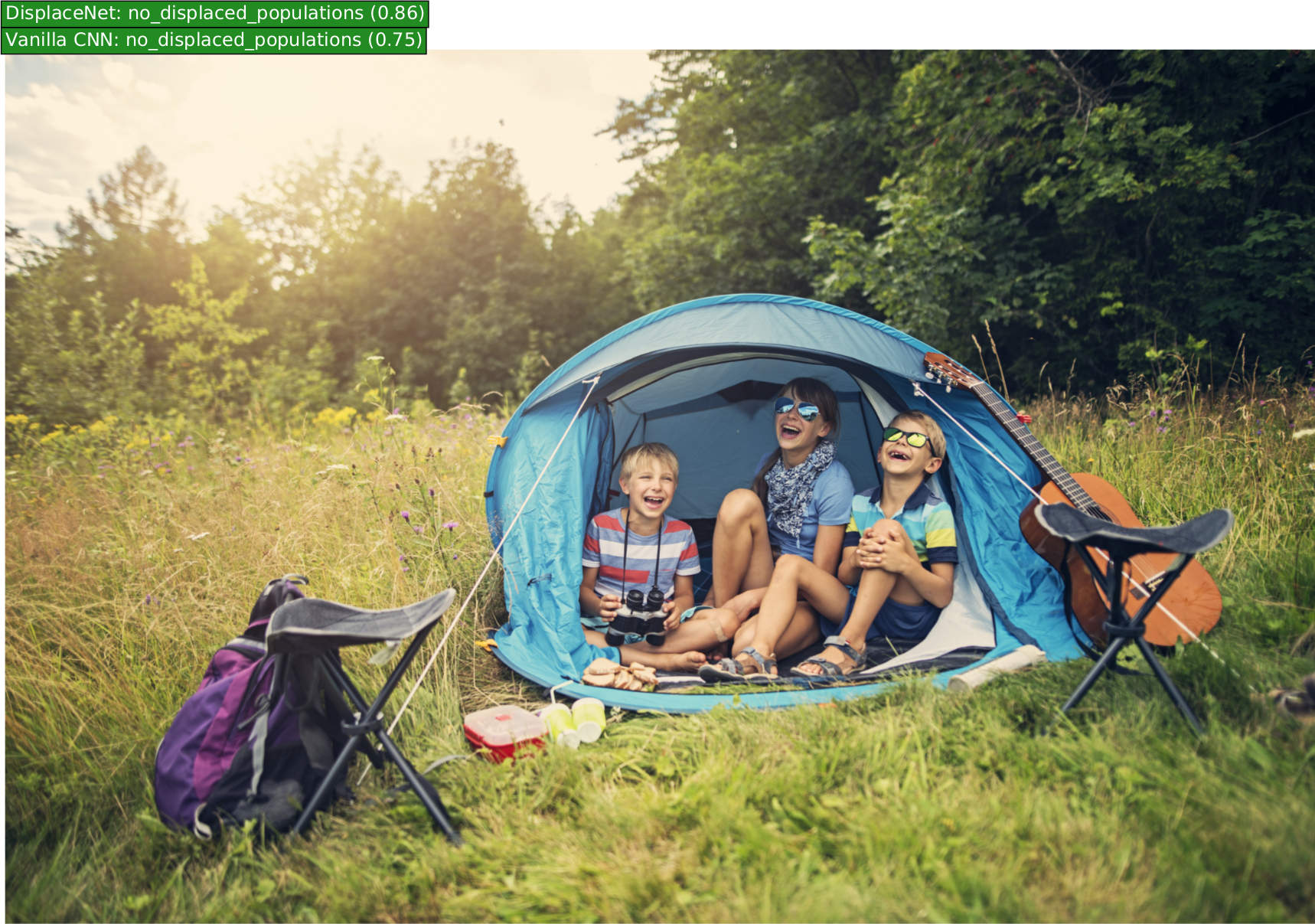} \\
		(e) & (f)  & (g)\\[1pt]
	\end{tabular}
	\caption{Displaced people detected by our method. Each image shows two predictions alongside their probabilities. Top prediction is given by DisplaceNet, while the bottom prediction is given by the respective vanilla CNN. Green colour implies that no displace people were detected, while red colour signifies that displaced people were detected. In some instances such as (a), (b), and (c), DisplaceNet overturns the initial-false prediction of the vanilla CNN, where in other instances such as (e), (f) and (g), DisplaceNet strengthens the initial-true prediction, resulting in higher coverage.}
	\label{Fig. 5}
\end{figure*}

\subsection{Training}

Due to different datasets, convergence times and loss imbalance, all three branches have been trained separately. For object detection we adopted an existing implementation of the RetinaNet object detector, pre-trained on the COCO dataset \cite{lin2014microsoft}, with a ResNet-50 backbone. 

For emotion recognition in continuous dimensions, we formulate this task as a regression problem using the Euclidean loss. The two feature extraction modules are designed as truncated versions of various well-known CNNs and initialised using pretrained models on two large-scale image classification datasets, ImageNet \cite{krizhevsky2012imagenet} and Places \cite{zhou2018places}. The truncated version of those CNNs removes the fully connected layer and outputs features from the last convolutional layer in order to maintain the localisation of different parts of the images which is significant for the task at hand. Features extracted from these two modules (red and blue boxes in Fig. \ref{Fig. 2}B) are then combined by a fusion module. This module first uses a global average pooling layer to reduce the number of features from each network and then a fully connected layer, with an output of a 256-D vector,  functions as a dimensionality reduction layer for the concatenated pooled features. Finally, we include a second fully connected layer with 3 neurons representing valence, arousal and dominance. The parameters of the three modules are learned jointly using stochastic gradient descent with momentum of 0.9. The batch size is set to 54 and we use dropout with a ratio of 0.5.

We formulate displaced people recognition as a binary classification problem. We train an end-to-end model for classifying everyday images as displaced people positive or displaced people negative, based on the context of the images. We fine-tune various CNN models for the two-class classification task. First, we conduct feature extraction utilising only the convolutional base of the original networks in order to end up with more generic representations as well as retaining spatial information similar to emotion recognition pipeline. The second phase consists of unfreezing some of the top layers of the convolutional base and jointly training a newly added fully connected layer and these top layers.

All the CNNs presented here were trained using the Keras Python deep learning framework \cite{chollet2015keras} over TensorFlow \cite{abadi2016tensorflow} on Nvidia GPU P100.

\section{Dataset and Metrics}

There are a limited number of image datasets for human rights violations recognition \cite{kalliatakis2019exploring, visapp17}. In order to find the main test platform on which we could demonstrate the effectiveness of DisplaceNet and analyse its various components, we construct a new image dataset by maintaining the verified samples intact for the category \textit{displaced populations}. The constructed dataset contains 609 images of displaced people and the same number of non displaced people counterparts for training, as well as 100 images collected from the web for testing and validation. The dataset is made publicly available for future research.
We evaluate DisplaceNet with two metrics \textit{accuracy} and \textit{coverage} and compare its performance against the sole use of a CNN classifier.

\section{Experiments}

\noindent
\textbf{Implementation details.} Our emotion recognition implementation is based on the emotion recognition in context (EMOTIC) model \cite{kosti2017emotion}, with the difference that our model estimates only continuous dimensions in VAD space. We train the three main modules on the EMOTIC database, which contains a total number of 18,316 images with 23,788 annotated people, using pre-trained CNN feature extraction modules. We treat this multiclass-multilabel problem as a regression problem by using a weighted Euclidean loss to compensate for the class imbalance of EMOTIC. 

For the classification part, we fine-tune our models for 50 iterations on the HRA subset with a learning rate of 0.0001 using the stochastic gradient descent (SGD) \cite{lecun1989backpropagation} optimizer for cross-entropy minimization. These \textit{vanilla} models will be examined against DisplaceNet. Here, vanilla means pure image classification using solely fine-tuning without any alteration.

\noindent
\textbf{Baseline.} To enable a fair comparison between vanilla CNNs and DisplaceNet, we use the same backbone combinations for all modules described in Fig. \ref{Fig. 2}. We report comparisons in both \textit{accuracy} and \textit{coverage} metrics for fine-tuning up to two convolutional layers in order to be consistent with the implementation of \cite{kalliatakis2019exploring}. The per-network results are shown in Table \ref{tab1}. The implementation of vanilla CNNs is solid with up to 61.5\% accuracy. 
Regarding coverage, vanilla CNNs achieve up to 16.83\%. This shows that it is possible to trade coverage with accuracy in the context of human rights image analysis. One can always obtain high accuracy by refusing to process a number of examples, but this reduces the coverage of the system. Nevertheless, vanilla CNNs provide a strong baseline to which we will compare our method.

Our method, has a mean coverage of 20.83\%. This is an absolute gain of 4 points over the baseline of 16.83\%. This is a relative improvement of 23.76\%. In relation to accuracy, DisplaceNet has a mean accuracy of 57.33\% which is an absolute drop of 4.17 points over the strong baselines of 61.5\%. This indicates a relative loss of only 6.7\%. We believe that this negligible drop in accuracy is mainly due to the fact that the $test$ set is not solely made up of images with people in their context, it also contains images of generic objects and scenes, where only the sole classifier's prediction is taken into account.

\begin{table}[t!]
	\centering
	\resizebox{\columnwidth}{!}{%
		\begin{tabular}{c|c|cc|cc}
			\multirow{2}{*}{\begin{tabular}[c]{@{}c@{}}backbone \\ network\end{tabular}} & \multirow{2}{*}{\begin{tabular}[c]{@{}c@{}}layers \\ fine-tuned\end{tabular}} & \multicolumn{2}{c|}{vanilla CNN} & \multicolumn{2}{c|}{\textbf{DisplaceNet}} \\ \cline{3-6} 
			&                                                                               & Top-1 acc.       & Coverage      & Top-1 acc.     & \textbf{Coverage}     \\ \hline
			VGG16                                                                        & \multirow{4}{*}{1}                                                            & 58\%             & 0\%           & 54\%           & \textbf{3\%}       \\
			VGG19                                                                        &                                                                               & 69\%             & 3\%           & 60\%           & \textbf{6\%}         \\
			ResNet50                                                                     &                                                                               & 60\%             & 0\%           & 55\%           & \textbf{4\%}         \\ \hline
			VGG16                                                                        & \multirow{4}{*}{2}                                                            & 63\%             & 43\%          & 63\%           & \textbf{49\%}         \\
			VGG19                                                                        &                                                                               & 77\%             & 54\%          & 74\%           & \textbf{58\%}         \\
			ResNet50                                                                     &                                                                               & 42\%             & 1\%           & 38\%           & \textbf{5\%}         \\ \hline
			\textbf{mean}                                                                        & -                                                            & 61.5\%             & 16.83\%            & 57.33\%           & \textbf{20.83\%}           \\ \hline
		\end{tabular}
	}
	\caption{\label{tab1}Detailed results on displaced people recognition using DisplaceNet. We show the main baseline and DisplaceNet for various network backbones. We bold the leading entries on coverage.}
\end{table}

\noindent
\textbf{Qualitative results.} We show our human rights abuse detection results in Fig. \ref{Fig. 5}. Each subplot illustrates two predictions alongside their probability scores. The top of the two predictions is given by DisplaceNet, while the bottom one is given by the respective vanilla CNN sole classifier. Our method can successfully classify displaced people by overturning the initial-false prediction of the vanilla CNN. Moreover, DisplaceNet can strengthen the initial-true prediction of the sole classifier. Finally, our method can be incorrect, because of false dominance score inferences. Some of them are caused by a failure of continuous dimensions emotion recognition, which is an interesting open problem for future research.

\section{Conclusion}

We have presented a human-centric approach for recognising displaced people from still images. This two-class labelling problem is not trivial, given the high-level image interpretation required. Understanding a person's control level of the situation from his frame of reference is closely related with situations where people have been forcibly displaced. Thus, the key to our computational framework is people's dominance level, which resonates well with our own common sense in judging potential displacement cases. We introduce the overall dominance level of the image which is responsible for weighting the classifiers prediction during inference. We benchmark performance of our DisplaceNet model against sole CNN classifiers. Our experimental results showed that this is an effective strategy, which we believe has good potential beyond human rights related classification. We hope this paper will spark interest and subsequent work along this line of research. All our code and data are publicly available.

%



%


{\small
\bibliographystyle{ieee_fullname}
\bibliography{egbib}

\begin{thebibliography}{10}\itemsep=-1pt

\bibitem{abadi2016tensorflow}
Mart{\'\i}n Abadi, Paul Barham, Jianmin Chen, Zhifeng Chen, Andy Davis, Jeffrey
  Dean, Matthieu Devin, Sanjay Ghemawat, Geoffrey Irving, Michael Isard, et~al.
\newblock Tensorflow: a system for large-scale machine learning.
\newblock In {\em OSDI}, volume~16, pages 265--283, 2016.

\bibitem{aronson2018computer}
Jay~D Aronson.
\newblock Computer vision and machine learning for human rights video analysis:
  Case studies, possibilities, concerns, and limitations.
\newblock {\em Law \& Social Inquiry}, 2018.

\bibitem{aronson2015video}
Jay~D Aronson, Shicheng Xu, and Alex Hauptmann.
\newblock Video analytics for conflict monitoring and human rights
  documentation.
\newblock {\em Center for Human Fights Science Technical Report}, 2015.

\bibitem{chollet2015keras}
Fran\c{c}ois Chollet et~al.
\newblock Keras.
\newblock \url{https://keras.io}, 2015.

\bibitem{ekman1971constants}
Paul Ekman and Wallace~V Friesen.
\newblock Constants across cultures in the face and emotion.
\newblock {\em Journal of personality and social psychology}, 17(2):124, 1971.

\bibitem{eleftheriadis2016joint}
Stefanos Eleftheriadis, Ognjen Rudovic, and Maja Pantic.
\newblock Joint facial action unit detection and feature fusion: A
  multi-conditional learning approach.
\newblock {\em IEEE transactions on image processing}, 25(12):5727--5742, 2016.

\bibitem{fabian2016emotionet}
C Fabian Benitez-Quiroz, Ramprakash Srinivasan, and Aleix~M Martinez.
\newblock Emotionet: An accurate, real-time algorithm for the automatic
  annotation of a million facial expressions in the wild.
\newblock In {\em Proceedings of the IEEE Conference on Computer Vision and
  Pattern Recognition}, pages 5562--5570, 2016.

\bibitem{global_trends_report}
UN~High~Commissioner for Refugees.
\newblock Global trends forced displacement in 2017.
\newblock Technical report, 2017.

\bibitem{Fu2017DSSDD}
Cheng-Yang Fu, Wei Liu, Ananth Ranga, Ambrish Tyagi, and Alexander~C. Berg.
\newblock Dssd : Deconvolutional single shot detector.
\newblock {\em CoRR}, abs/1701.06659, 2017.

\bibitem{girshick2015fast}
Ross Girshick.
\newblock Fast r-cnn.
\newblock In {\em Proceedings of the IEEE international conference on computer
  vision}, pages 1440--1448, 2015.

\bibitem{girshick2014rich}
Ross Girshick, Jeff Donahue, Trevor Darrell, and Jitendra Malik.
\newblock Rich feature hierarchies for accurate object detection and semantic
  segmentation.
\newblock In {\em Proceedings of the IEEE conference on computer vision and
  pattern recognition}, pages 580--587, 2014.

\bibitem{he2016deep}
Kaiming He, Xiangyu Zhang, Shaoqing Ren, and Jian Sun.
\newblock Deep residual learning for image recognition.
\newblock In {\em Proceedings of the IEEE conference on computer vision and
  pattern recognition}, pages 770--778, 2016.

\bibitem{visapp17}
Grigorios Kalliatakis, Shoaib Ehsan, Maria Fasli, Ales Leonardis, Juergen Gall,
  and Klaus~D. McDonald-Maier.
\newblock Detection of human rights violations in images: Can convolutional
  neural networks help?
\newblock In {\em Proceedings of the 12th International Joint Conference on
  Computer Vision, Imaging and Computer Graphics Theory and Applications -
  Volume 5: VISAPP, (VISIGRAPP 2017)}, pages 289--296. SciTePress, 2017.

\bibitem{kalliatakis2019exploring}
Grigorios Kalliatakis, Shoaib Ehsan, Ale{\v{s}} Leonardis, Maria Fasli, and
  Klaus~D McDonald-Maier.
\newblock Exploring object-centric and scene-centric cnn features and their
  complementarity for human rights violations recognition in images.
\newblock {\em IEEE Access}, 7:10045--10056, 2019.

\bibitem{kosti2017emotion}
Ronak Kosti, Jose~M Alvarez, Adria Recasens, and Agata Lapedriza.
\newblock Emotion recognition in context.
\newblock In {\em The IEEE Conference on Computer Vision and Pattern
  Recognition (CVPR)}, volume~1, 2017.

\bibitem{krizhevsky2012imagenet}
Alex Krizhevsky, Ilya Sutskever, and Geoffrey~E Hinton.
\newblock Imagenet classification with deep convolutional neural networks.
\newblock In {\em Advances in neural information processing systems}, pages
  1097--1105, 2012.

\bibitem{lecun1989backpropagation}
Yann LeCun, Bernhard Boser, John~S Denker, Donnie Henderson, Richard~E Howard,
  Wayne Hubbard, and Lawrence~D Jackel.
\newblock Backpropagation applied to handwritten zip code recognition.
\newblock {\em Neural computation}, 1(4):541--551, 1989.

\bibitem{lin2018focal}
Tsung-Yi Lin, Priyal Goyal, Ross Girshick, Kaiming He, and Piotr Doll{\'a}r.
\newblock Focal loss for dense object detection.
\newblock {\em IEEE transactions on pattern analysis and machine intelligence},
  2018.

\bibitem{lin2014microsoft}
Tsung-Yi Lin, Michael Maire, Serge Belongie, James Hays, Pietro Perona, Deva
  Ramanan, Piotr Doll{\'a}r, and C~Lawrence Zitnick.
\newblock Microsoft coco: Common objects in context.
\newblock In {\em European conference on computer vision}, pages 740--755.
  Springer, 2014.

\bibitem{liu2016ssd}
Wei Liu, Dragomir Anguelov, Dumitru Erhan, Christian Szegedy, Scott Reed,
  Cheng-Yang Fu, and Alexander~C Berg.
\newblock Ssd: Single shot multibox detector.
\newblock In {\em European conference on computer vision}, pages 21--37.
  Springer, 2016.

\bibitem{mehrabian1995framework}
Albert Mehrabian.
\newblock Framework for a comprehensive description and measurement of
  emotional states.
\newblock {\em Genetic, social, and general psychology monographs}, 1995.

\bibitem{mehrabian1974approach}
Albert Mehrabian and James~A Russell.
\newblock {\em An approach to environmental psychology.}
\newblock the MIT Press, 1974.

\bibitem{piracés_2018}
Enrique Piracés.
\newblock {\em The Future of Human Rights Technology}, page 289–308.
\newblock Cambridge University Press, 2018.

\bibitem{redmon2016you}
Joseph Redmon, Santosh Divvala, Ross Girshick, and Ali Farhadi.
\newblock You only look once: Unified, real-time object detection.
\newblock In {\em Proceedings of the IEEE conference on computer vision and
  pattern recognition}, pages 779--788, 2016.

\bibitem{redmon2016yolo9000}
Joseph Redmon and Ali Farhadi.
\newblock Yolo9000: Better, faster, stronger.
\newblock {\em 2017 IEEE Conference on Computer Vision and Pattern Recognition
  (CVPR)}, pages 6517--6525, 2017.

\bibitem{ren2015faster}
Shaoqing Ren, Kaiming He, Ross Girshick, and Jian Sun.
\newblock Faster r-cnn: Towards real-time object detection with region proposal
  networks.
\newblock In {\em Advances in neural information processing systems}, pages
  91--99, 2015.

\bibitem{OverFeat}
Pierre Sermanet, David Eigen, Xiang Zhang, Micha{\"{e}}l Mathieu, Rob Fergus,
  and Yann LeCun.
\newblock Overfeat: Integrated recognition, localization and detection using
  convolutional networks.
\newblock {\em International Conference on Learning Representations (ICLR)},
  2014.

\bibitem{uijlings2013selective}
Jasper~RR Uijlings, Koen~EA Van De~Sande, Theo Gevers, and Arnold~WM Smeulders.
\newblock Selective search for object recognition.
\newblock {\em International journal of computer vision}, 104(2):154--171,
  2013.

\bibitem{zhou2018places}
Bolei Zhou, Agata Lapedriza, Aditya Khosla, Aude Oliva, and Antonio Torralba.
\newblock Places: A 10 million image database for scene recognition.
\newblock {\em IEEE transactions on pattern analysis and machine intelligence},
  40(6):1452--1464, 2018.

\end{thebibliography}
}

\end{document}